\documentclass[letterpaper]{article} % DO NOT CHANGE THIS
\usepackage{aaai2026}  % DO NOT CHANGE THIS
\usepackage{times}  % DO NOT CHANGE THIS
\usepackage{helvet}  % DO NOT CHANGE THIS
\usepackage{courier}  % DO NOT CHANGE THIS
\usepackage[hyphens]{url}  % DO NOT CHANGE THIS
\usepackage{graphicx} % DO NOT CHANGE THIS
\usepackage{natbib}  % DO NOT CHANGE THIS AND DO NOT ADD ANY OPTIONS TO IT
\usepackage{caption} % DO NOT CHANGE THIS AND DO NOT ADD ANY OPTIONS TO IT
\usepackage{algorithm}
\usepackage{algorithmic}

\usepackage{amssymb}
\usepackage{booktabs}
 \usepackage{amsmath}
 \usepackage{placeins}
\usepackage{newfloat}
\usepackage{listings}
\DeclareCaptionStyle{ruled}{labelfont=normalfont,labelsep=colon,strut=off} % DO NOT CHANGE THIS
\floatstyle{ruled}
\newfloat{listing}{tb}{lst}{}
\floatname{listing}{Listing}
\title{RealRep: Generalized SDR-to-HDR Conversion via Attribute-Disentangled Representation Learning}

\author{
    Li Xu\textsuperscript{\rm 1}\equalcontrib,
    Siqi Wang\textsuperscript{\rm 1}\textsuperscript{$\dagger$}\equalcontrib,
    Kepeng Xu\textsuperscript{\rm 1}\textsuperscript{$\dagger$},
    Lin Zhang\textsuperscript{\rm 1},
    Gang He\textsuperscript{\rm 1},
    Weiran Wang\textsuperscript{\rm 1},
    Yu-Wing Tai\textsuperscript{\rm 2}
}

\affiliations{
    \textsuperscript{\rm 1}Xidian University \\
    \textsuperscript{\rm 2}Dartmouth College 
}

\usepackage{bibentry}
\begin{document}

\maketitle

{ % 使用花括号隔离，防止影响后续的正常脚注
  \makeatletter
  \def\@makefnmark{} % 核心：定义“脚注标记”为空，这样 "0" 就不会被打印出来
  \makeatother
  
  % 在这里放入你的脚注文本
  \footnotetext{$\dagger$ Corresponding authors.} 
}

\begin{abstract}

High-Dynamic-Range Wide-Color-Gamut (HDR-WCG) technology is becoming increasingly widespread, driving a growing need for converting Standard Dynamic Range (SDR) content to HDR. Existing methods primarily rely on fixed tone mapping operators, which struggle to handle the diverse appearances and degradations commonly present in real-world SDR content.
To address this limitation, we propose a generalized SDR-to-HDR framework that enhances robustness by learning attribute-disentangled representations. Central to our approach is Realistic Attribute-Disentangled Representation Learning (RealRep), which explicitly disentangles luminance and chrominance components to capture intrinsic content variations across different SDR distributions. Furthermore, we design a Luma-/Chroma-aware negative exemplar generation strategy that constructs degradation-sensitive contrastive pairs, effectively modeling tone discrepancies across SDR styles.
Building on these attribute-level priors, we introduce the Degradation-Domain Aware Controlled Mapping Network (DDACMNet), a lightweight, two-stage framework that performs adaptive hierarchical mapping guided by a control-aware normalization mechanism. DDACMNet dynamically modulates the mapping process via degradation-conditioned features, enabling robust adaptation across diverse degradation domains.
Extensive experiments demonstrate that RealRep consistently outperforms state-of-the-art methods in both generalization and perceptually faithful HDR color gamut reconstruction.
\end{abstract}

% \vspace{-2mm}
\section{Introduction}
% {\fontsize{7}{12}\selectfont
% High Dynamic Range/Wide Color Gamut (HDR/WCG) media
% }

High Dynamic Range/Wide Color Gamut (HDR/WCG) media expands luminance range and visible colors beyond Standard Dynamic Range (SDR) limitations, enhancing visual experiences \cite{BT2020,SMPTE2014}. Advances in display technologies, such as PQ and HLG Electro-Optical Transfer Functions (OETFs) \cite{ITU-R_BT2100-1} and BT.2020 Wide Color Gamut (WCG) primaries \cite{ITU-R_BT2020-2}, have further elevated HDR's benefits. However, most existing media content remains in SDR format, limiting its potential on HDR displays. Inverse Tone Mapping (iTM), converting SDR to HDR, is thus essential for fully utilizing HDR technology and repurposing SDR content.

\begin{figure}[!t]
	\centering
	\includegraphics[width=0.47\textwidth]{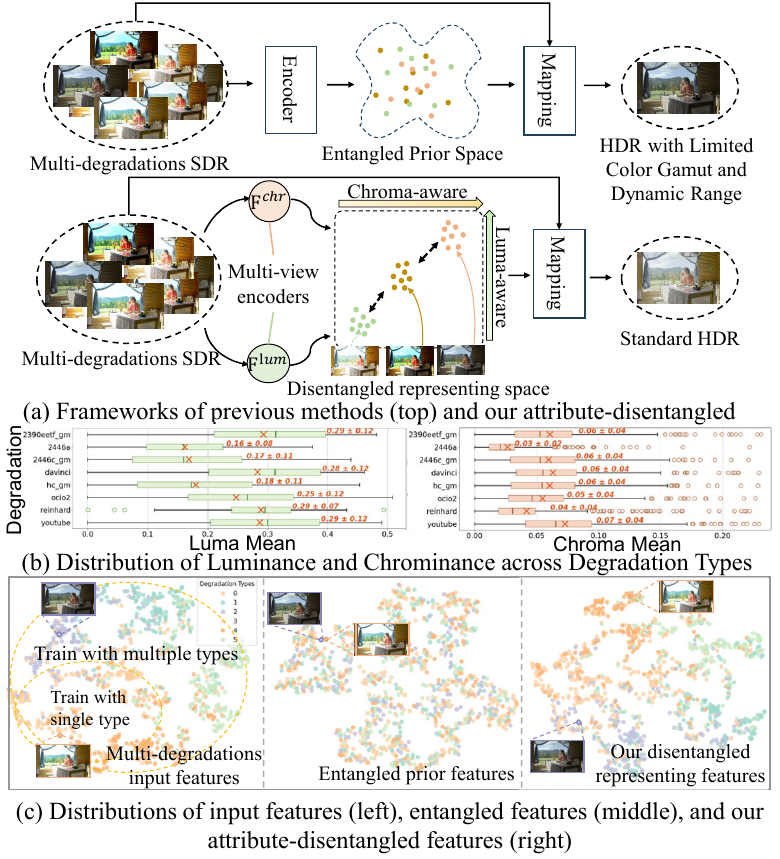}
	\caption{An illustration of our motivation. (a) Comparison between previous frameworks (top) and our attribute-disentangled method (bottom), which explicitly separates luminance and chrominance and injects them into the learned prior space. (b) Distribution shifts of luminance and chrominance across degradations, motivating the need for disentangled modeling to ensure robustness. (c) t-SNE visualization of SDR input features (left), features from previous methods (middle), and our method (right), all trained on a multi-degradation dataset. Our approach achieves superior attribute separation, enabling better generalization across diverse SDR conditions.} 
    % \vspace{-4mm}
    \label{Figure 1}
\end{figure}

\begin{figure*}[t]
	\centering  %图片全局居中
	\includegraphics[width=0.999\textwidth]{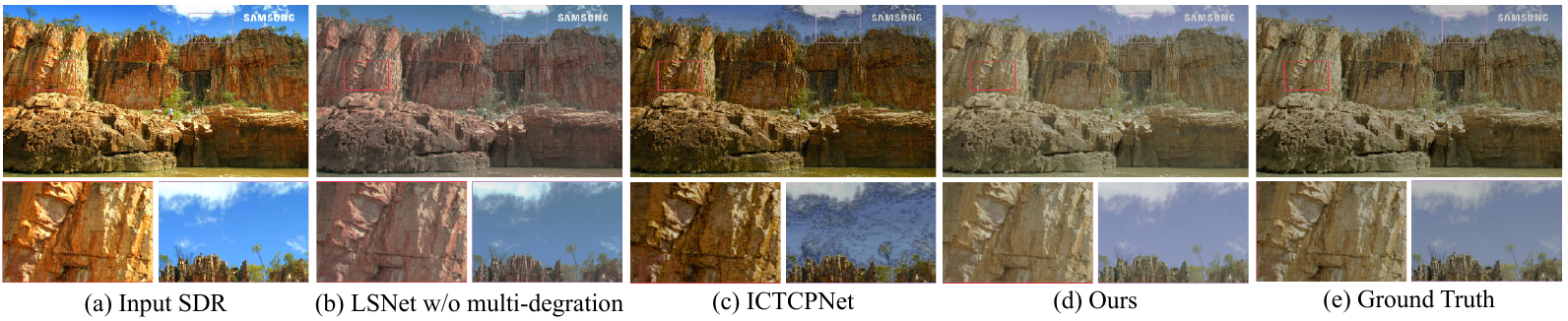}
	\caption{The superiority of our approach in recovering luminance and chrominance under unknown degradations. (a) Input SDR image. (b) Results of LSNet \protect\cite{Guo2023}, trained on a single degradation type, produces results that closely mimic the input SDR. (c) Results of ICTCPNet \protect\cite{Huang2023VideoIT} with multi-degradation training, which still fail to recover realistic brightness and color. (d) Our proposed method (RealRep) achieves significantly better recovery of both luminance and chrominance, demonstrating strong generalization to unseen degradations. (e) Ground truth HDR image.} 
    \label{Figure 2}
    % \vspace{-2mm}
\end{figure*}

Early iTM methods relied on heuristic tone-mapping rules \cite{Kim2018,Kim2019Deep,Kim2019JSIGANGJ}, which struggled with over-/under-exposed regions and lacked generality. As deep learning evolves\cite{MaIR}, recent deep learning approaches \cite{Chen2021,Shao2022HybridCD,Xu2022FMNet,Guo2023,he2022sdrtv,xu2022sdrtv,xu2023towards,xu2024beyond,xu2024beyond2} treat iTM as a mapping problem and have improved visual fidelity. However, most are trained on SDRs synthesized by fixed tone-mapping (TM) curves, such as Reinhard or YouTube styles, and struggle to generalize to real-world SDRs with diverse degradations. As shown in Figure~\ref{Figure 1}(b), different degradation types exhibit distinct statistical variations in luminance and chrominance. Prior works often overlook chroma variations and learn entangled representations, resulting in poor separation of inputs with different styles (Figure~\ref{Figure 1}(c, middle)). In contrast, our method disentangles luminance and chrominance, learning attribute-aware representations that are clearly separated by degradation type (Figure~\ref{Figure 1}(c, right)), enabling better generalization to realistic SDR scenarios.

To address these challenges, we propose RealRep, a novel SDR-to-HDR framework designed for real-world scenarios involving complex and diverse degradations. RealRep employs an attribute-disentangled representation learning strategy that separately encode luminance and chrominance, allowing the model to capture degradation-aware, generalizable priors for inverse tone mapping. To further enhance mapping robustness, we introduce DDACMNet, a degradation-domain-aware controlled mapping network that leverages these priors through a control-aware normalization mechanism. By learning from a wide spectrum of degradation types, RealRep constructs a robust representation space that models both large-scale brightness variations and fine-grained color structures. As shown in Figure~\ref{Figure 2}, our method produces perceptually more faithful HDR reconstructions compared to state-of-the-art approaches, particularly under severe degradation and color distortion.

Our main contributions are summarized as follows:

\begin{itemize}
    \item \textbf{Attribute-disentangled representation learning:} We propose RealRep, which disentangles luminance and chrominance to reduce style variation and builds a robust embedding space for accurate SDR-to-HDR conversion.
    
    \item \textbf{Degradation-controlled mapping:} We design DDACMNet to inject disentangled priors via a zero-initialized controller and stabilize early training through a two-stage degradation-guided strategy.
    
    \item \textbf{Robust performance on diverse SDR styles:} Extensive experiments on datasets with varied and unseen degradations show that RealRep achieves superior visual quality and generalization over state-of-the-art methods.
\end{itemize}

\begin{figure*}
	\centering  %图片全局居中
	\includegraphics[width=0.999\textwidth]{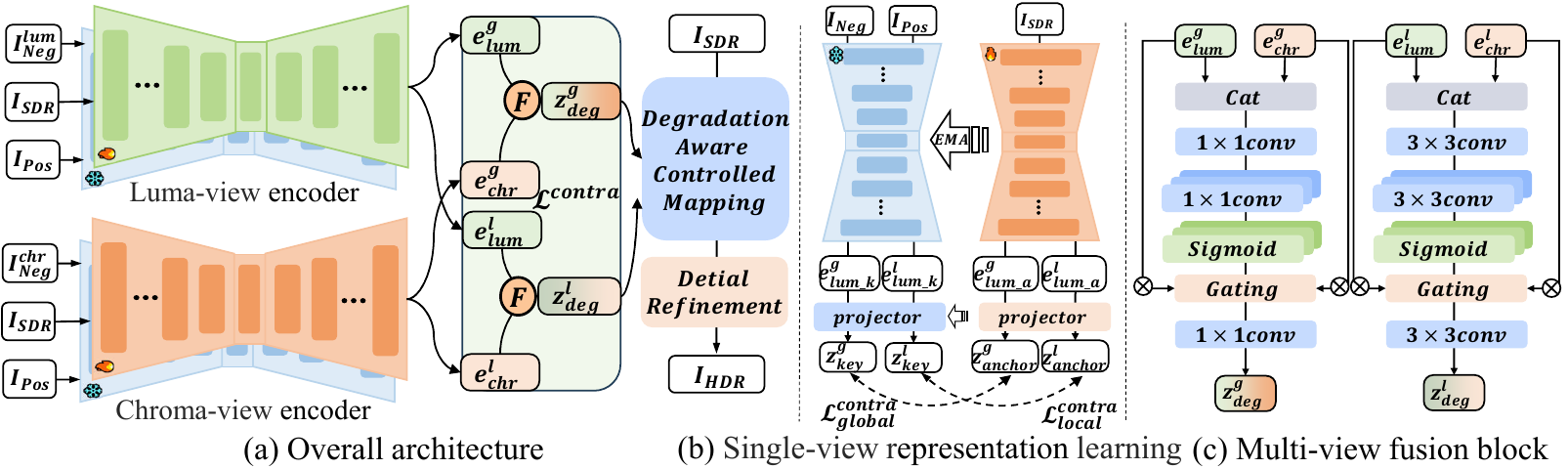}
       % % \vspace{-3mm}
	\caption{Overview of the proposed multi-degradation SDR-to-HDR framework. Our architecture consists of a pair of multi-view degradation encoders that separately extract global and local representations of luminance and chrominance, a multi-view fusion module that facilitates spatial and channel-wise interaction, and the Degradation-Domain Aware Controlled Mapping Network (DDACMNet), which performs adaptive mapping and detail refinement conditioned on degradation-aware features. The encoders are jointly trained with a disentanglement objective to ensure cross-domain consistency. Our framework enables dynamic luminance expansion and perceptually faithful color gamut reconstruction across a wide range of real-world SDR degradations.} 
    \label{Figure 3}
    % \vspace{-4mm}
\end{figure*}

\section{Related Work}
\subsection{Multiple Styles SDR-to-HDR}

SDR-to-HDR conversion has recently attracted growing attention due to its relevance in media enhancement and consumer displays. Kim et al. \cite{Kim2018} pioneered a joint framework for super-resolution and SDR-to-HDR conversion using CNNs. Deep SR-ITM \cite{Kim2019Deep} introduced spatially adaptive contrast modulation, while JSI-GAN \cite{Kim2019JSIGANGJ} incorporated dynamic convolution and adversarial learning to enhance visual quality.
To support learning-based methods, HDRTV1K \cite{Chen2021} and HDRTV4K \cite{Guo2023} were proposed, along with HDRTVNet and LSNet. These models combine global color mapping, local enhancement, and luminance-aware strategies. FMNet \cite{Xu2022FMNet} further introduced frequency-aware modulation to reduce artifacts.
Despite these advances, most approaches rely on synthetic SDRs generated by fixed tone-mapping curves (e.g., Reinhard, Hable), which limits generalization to real-world SDR types. Guo et al. \cite{Guo2023} also emphasized this issue.
Our method addresses this limitation by learning attribute-disentangled representations from diverse SDR types, enabling robust luminance and chrominance modeling without reliance on predefined tone-mapping operators.

\subsection{Image Enhancement for Multi-Degradations}

Image enhancement tasks, such as super-resolution, denoising, dehazing, deraining, and deblurring, were traditionally addressed separately. Early CNN-based methods like SRCNN~\cite{Dong_Loy_He_Tang_2014} and its deeper extensions~\cite{Kim_Lee_Lee_2016,Lim_Son_Kim_Nah_Lee_2017,Ahn_Kang_Sohn_2018,Ma2023Overview} achieved strong results in super-resolution, while similar task-specific networks were developed for other degradations~\cite{Tian_Xu_Li_Zuo_Fei_Liu_2020,Li_Peng_Wang_Xu_Feng_2017,Yang_Liu_Yang_Guo_2019,Eboli_Sun_Ponce_2020}.
To improve generalization, recent works explore unified models. Transformer-based frameworks like IPT~\cite{Chen_Wang2021}, AirNet~\cite{Li2022AllInOneIR}, and PromptIR~\cite{Potlapalli2023PromptIRPF} handle multiple degradations within a single network.
While unified models improve scalability under unknown degradations, they offer limited control in HDR contexts. Our method targets HDR enhancement by disentangling luminance and chrominance, enabling more robust and interpretable reconstruction.

\subsection{Degradation Representation}

Degradation representation is critical for guiding effective enhancement, especially under diverse or unknown degradation conditions. Recent approaches learn implicit representations that encode degradation characteristics without explicit labels. AirNet~\cite{Li2022AllInOneIR} leverages contrastive learning, while PromptIR~\cite{Potlapalli2023PromptIRPF} conditions restoration with learned prompts. We adopt a similar strategy for multi-style SDR-to-HDR enhancement, learning a degradation-aware representation that captures style-specific information to guide adaptive HDR reconstruction.

\section{Methodology}

\subsection{Overview} 

Figure 3 shows an overview of our RealRep framework. It features:  
(1) a contrastive multi-view encoder that disentangles luminance and chrominance across global and local views, and  
(2) a degradation-aware mapping network (DDACMNet) guided by hierarchical priors.

Given an SDR input \( x \), the encoder extracts luminance and chrominance features from global and local views. These are projected and normalized into a multi-view representation \( z \), and supervised via contrastive loss to encourage degradation-invariant and attribute-disentangled learning. A fusion block aggregates \( z \) into a unified latent code.

DDACMNet performs progressive SDR-to-HDR mapping, modulated by global and local degradation priors derived from the encoder. This enables both global consistency and spatial adaptivity to diverse SDR degradations.

The overall process is defined as:
\begin{equation}
    \hat{y} = f_{\text{DDACM}}\left(f_{\text{fusion}}\left(e_{\text{lum}}^{g,l}, e_{\text{chr}}^{g,l}\right),\; x\right).
\end{equation}

\subsection{Attribute-Disentangled Representation Learning}

\subsubsection{Attribute Feature Extraction.}  
We adopt a UNet-based encoder~\cite{Ronneberger2015UNetCN} to extract hierarchical features from SDR inputs. The encoder-decoder structure, enhanced with skip connections, is well-suited for SDR-to-HDR tasks that require both global tone mapping and local detail preservation~\cite{Chen2021}. 

Given an input SDR image \( x \), we extract two modality-specific features: the luminance feature \( e_{\text{lum}} \in \mathbb{R}^{H \times W \times C} \) and the chrominance feature \( e_{\text{chr}} \in \mathbb{R}^{H \times W \times C} \). Each feature is further decomposed into global and local views. Specifically, the global features \( e_c^g \) are extracted from the UNet bottleneck via a \( 1 \times 1 \) convolution followed by global average pooling, while the local features \( e_c^l \) are obtained from the decoder and fused with encoder outputs through skip connections, where \( c \in \{\text{lum}, \text{chr}\} \). Each view-specific feature is then passed through a projection head \( \phi_c^s \) and normalized using \( \ell_2 \)-normalization, resulting in the following embeddings:

\begin{equation}
z_c^s = \frac{\phi_c^s(e_c^s)}{\left\|\phi_c^s(e_c^s)\right\|_2}, \quad \text{for } c \in \{\text{lum}, \text{chr}\},\ s \in \{g, l\}.
\end{equation}

These four embeddings \( \{z_{\text{lum}}^g, z_{\text{lum}}^l, z_{\text{chr}}^g, z_{\text{chr}}^l\} \) form the basis of our multi-view representation, and are later supervised using contrastive learning to enforce attribute-level disentanglement.

\begin{algorithm}[tb]
    \caption{Luma-/Chroma-aware Negative Exemplars Generation}
    \label{alg:negative_samples_simplified_en}
    \textbf{Input}: Multi-degradation dataset $D$, parameters $k_l$ (luminance negatives), $k_c$ (chrominance negatives)\\
    \textbf{Output}: Luminance negatives $N_l$, Chrominance negatives $N_c$
    \begin{algorithmic}[1]
        \STATE Initialize $N_l \leftarrow []$, $N_c \leftarrow []$
        
        \FOR{each image $I \in D$}
            \STATE Convert to YCbCr: Extract brightness ($L_{\text{ref}}$) and color components ($C_{\text{ref}}$)
            \STATE Initialize lists $List_l \leftarrow []$, $List_c \leftarrow []$
            
            \FOR{each degradation $x_d \in I.\text{degradations}$}
                \STATE Extract brightness ($L_d$) and color components ($C_d$) of $x_d$
                \STATE Compute differences: $D_l \leftarrow \|L_d - L_{\text{ref}}\|_1$, $D_c \leftarrow \|C_d - C_{\text{ref}}\|_1$
                \STATE Append $(D_l, x_d)$ to $List_l$, $(D_c, x_d)$ to $List_c$
            \ENDFOR
            
            \STATE Select top $k_l$ luminance-based and top $k_c$ chrominance-based samples
            \STATE $I_l \leftarrow \text{TopK}(List_l, k_l)$, $I_c \leftarrow \text{TopK}(List_c, k_c)$
            
            \FOR{each $(x_l, x_c)$ in $I_l$ and $I_c$}
                \STATE Add $(L_l, C_{\text{ref}})$ to $N_l$
                \STATE Add $(L_{\text{ref}}, C_c)$ to $N_c$
            \ENDFOR
        \ENDFOR
        
        \STATE Convert $N_l$, $N_c$ back to RGB
        \STATE \textbf{Return} $N_l$, $N_c$
    \end{algorithmic}
\end{algorithm}

\subsubsection{Disentangled Representation Learning.}
To learn degradation-invariant and attribute-disentangled representations for SDR-to-HDR mapping, we introduce a contrastive representation separation paradigm that explicitly leverages the statistical inconsistency between luminance and chrominance under different SDR degradations.

While SDR images derived from the same HDR source should share consistent scene semantics, they often exhibit significant variations in local brightness and color distributions due to diverse tone mapping processes. We exploit this observation by constructing contrastive samples that isolate such variations. Specifically, positive samples are generated by applying spatial augmentations (e.g., flips) to the same SDR input, which preserve both luminance and chrominance identity while introducing geometric diversity.

To construct hard negatives, we develop a Luma-/Chroma-aware mining strategy based on the assumption that SDR degradations primarily distort either brightness or color in a content-independent way. For each anchor, we compute the L1 distance over the luminance or chrominance channels with respect to a candidate pool, and select the top-\(k\) dissimilar samples as negatives. To further disentangle the contrastive supervision, we synthesize attribute-specific hard negatives by replacing either the luminance or chrominance channel of the anchor with that of a distinct SDR sample. This ensures that the contrastive signal focuses on a single modality at a time.

Formally, the contrastive loss is defined over all attribute-view combinations:
\begin{equation}
\resizebox{0.9\linewidth}{!}{$
\mathcal{L}_{\text{contra}} = \sum_{c \in \{\text{lum}, \text{chr}\}} \sum_{s \in \{\text{g}, \text{l}\}} -\log\left(\frac{e^{z_{c}^{s} \cdot z_{c}^{s+}}}{e^{z_{c}^{s} \cdot z_{c}^{s+}} + \sum_{j=1}^{N_{c}} e^{z_{c}^{s} \cdot z_{c,j}^{s}}}\right)
$}
\end{equation}
\noindent
where \(c \in \{\text{lum}, \text{chr}\}\) denotes the feature type, i.e., luminance (\(\text{lum}\)) or chrominance (\(\text{chr}\)); 
\(s \in \{\text{g}, \text{l}\}\) denotes the feature scope, i.e., global (\(\text{g}\)) or local (\(\text{l}\)); 
\(z_{c}^{s}\) is the feature vector of the anchor sample for feature type \(c\) and scope \(s\); 
\(z_{c}^{s+}\) is the positive sample, which is generated by applying data augmentation to the anchor sample \(z_{c}^{s}\); 
\(z_{c,j}^{s}\) are negative samples, where \(j = 1, 2, \dots, N_{c}\), and \(N_{c}\) is the total number of negative samples for feature type \(c\).

This objective encourages the model to align samples that share semantic content while explicitly penalizing variations in luminance and chrominance caused by SDR degradation. As a result, the learned embedding space becomes both semantically consistent and attribute-disentangled, providing a robust foundation for high-fidelity HDR reconstruction.

% \vspace{0.5em}
\subsubsection{Multi-View Attribute Fusion.}
After contrastive training, we fuse the view-specific embeddings into a unified latent representation. The fusion module leverages multi-head convolutions and learnable gating mechanisms to dynamically aggregate global-local and luminance-chrominance cues. The output embedding \( z \) integrates attribute-aware priors and is passed to the degradation-adaptive HDR mapping network described in Section~\ref{Figure 3}(c).

\begin{figure}[t]
	\centering
	\includegraphics[width=0.47\textwidth]{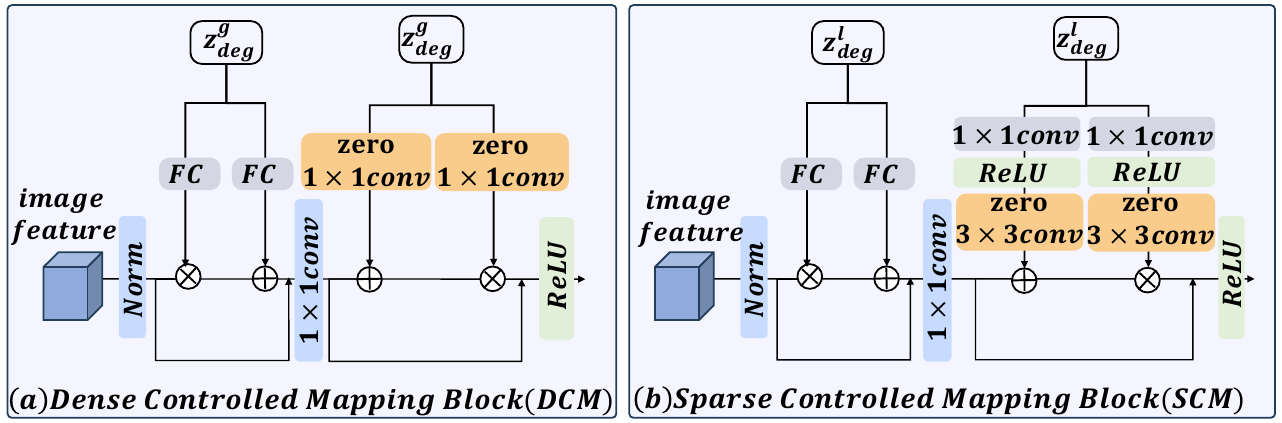}
	\caption{Illustration of (a) Dense Controlled Mapping (DCM) and (b) Sparse Controlled Mapping (SCM).} 
    % \vspace{-4mm}
    \label{controlled_mapping}
\end{figure}

\begin{table*}[h]
\centering
\setlength{\tabcolsep}{2.5pt} % 调整列间距
% \footnotesize % 调整字体大小为适度的比例
\normalsize

\resizebox{\linewidth}{!}{% 将宽度限制为页面宽度
\begin{tabular}{l|cccccccc|c}
\toprule
\textbf{Methods} &
  \textbf{2390eetf\_gm} &
  \textbf{2446a} &
  \textbf{2446c\_gm} &
  \textbf{davinci} &
  \textbf{hc\_gm} &
  \textbf{ocio2} &
  \textbf{reinhard} &
  \textbf{youtube} &
  % \textbf{hdrtv1k} &
  \textbf{Average} \\ 
  & \textbf{PSNR$\uparrow$ / SSIM$\uparrow$} &
  \textbf{PSNR$\uparrow$ / SSIM$\uparrow$} &
  \textbf{PSNR$\uparrow$ / SSIM$\uparrow$} &
  \textbf{PSNR$\uparrow$ / SSIM$\uparrow$} &
  \textbf{PSNR$\uparrow$ / SSIM$\uparrow$} &
  \textbf{PSNR$\uparrow$ / SSIM$\uparrow$} &
  \textbf{PSNR$\uparrow$ / SSIM$\uparrow$} &
  \textbf{PSNR$\uparrow$ / SSIM$\uparrow$} &
  \textbf{PSNR$\uparrow$ / SSIM$\uparrow$} \\
\midrule
HDRUNet &
  29.91 / 0.9219 &
  25.55 / 0.8861 &
  24.17 / 0.8509 &
  23.17 / 0.8790 &
  24.67 / 0.8548 &
  25.12 / 0.8676 &
  23.66 / 0.8887 &
  22.76 / 0.8758 &
  % 20.83 / 0.8667 &
  % 24.43 / 0.8768 \\
  24.88 / 0.8781\\
  % 24.88 / 0.8781\\
HDRTVNet &
  28.97 / 0.9059 &
  23.73 / 0.8502 &
  26.05 / 0.8597 &
  25.50 / 0.8815 &
  26.91 / 0.8621 &
  26.46 / 0.8699 &
  23.23 / 0.8722 &
  25.31 / 0.8709 &
  % \underline{23.78} / \underline{0.8824} &
  25.77 / 0.8716 \\
  % 25.55 / 0.8728 \\ 
FMNet &
  27.96 / 0.9050 &
  23.84 / 0.8753 &
  22.29 / 0.8327 &
  24.48 / 0.8835 &
  23.16 / 0.8390 &
  24.88 / 0.8603 &
  22.26 / 0.8603 &
  24.31 / 0.8755 &
  % 21.61 / 0.8609 &
  24.15 / 0.8665 \\
  % 23.87 / 0.8658 \\ 
ICTCPNet &
  29.58 / 0.8548 &
  27.02 / 0.8548 &
  27.76 / 0.8594 &
  25.67 / 0.8705 &
  \underline{28.27} / 0.8612 &
  \underline{29.97} / 0.8709 &
  25.63 / 0.8694 &
  25.71 / 0.8639 &
  % 21.61 / 0.8609 &
  27.45 / 0.8631 \\ 
LSNet &
  \underline{33.23} / \underline{0.9408} &
  \underline{31.18} / \underline{0.9080} &
  28.05 / \underline{0.8731} &
  25.63 / 0.9000 &
  27.34 / 0.8712 &
  29.56 / \underline{0.8913} &
  \underline{25.67} / \underline{0.8970} &
  27.01 / 0.9019 &
  % 20.36 / 0.8603 &
  28.46 / \underline{0.8979} \\
  % \underline{27.56} / \underline{0.8937} \\ 
\midrule
AirNet &
  23.00 / 0.8594 &
  21.72 / 0.8519 &
  21.02 / 0.8044 &
  22.22 / 0.8672	 &
  21.59 / 0.8107 &
  22.18 / 0.8407 &
  18.72 / 0.8379 &
  22.20 / 0.8512 &
  % 21.24 / 0.7946 &
  21.58 / 0.8404 \\
  % 21.58 / 0.8404 \\ 
PromptIR &
  30.06 / 0.9148 &
  30.72 / 0.9073 &
  \underline{28.23} / 0.8691 &
  27.83 / 0.9028 &
  28.00 / 0.8740 &
  29.30 / 0.8838 &
  24.97 / 0.8931 &
  27.62 / 0.9068 &
  % 21.24 / 0.7946 &
  28.34 / 0.8940 \\
  % 27.54 / 0.8799 \\ 
RAM-PromptIR &
  30.25 / 0.9188 &
  30.83 / 0.9103 &
  28.20 / 0.8711 &
  \underline{28.07} / \underline{0.9088} &
  28.09 / \underline{0.8770} &
  29.31 / 0.8828 &
  25.29 / 0.9001 &
  \underline{27.77} / \underline{0.9108} &
  \underline{28.48} / 0.8975 \\
\midrule
Ours &
  \textbf{34.13} / \textbf{0.9507} &
  \textbf{34.41} / \textbf{0.9333} &
  \textbf{30.38} / \textbf{0.8936} &
  \textbf{30.05} / \textbf{0.9266} &
  \textbf{30.47} / \textbf{0.8976} &
  \textbf{31.36} / \textbf{0.9083} &
  \textbf{27.59} / \textbf{0.9292} &
  \textbf{30.03} / \textbf{0.9358} &
  % \textbf{29.12} / \textbf{0.9412} &
  \textbf{31.05} / \textbf{0.9219} \\
  % \textbf{30.84} / \textbf{0.9240} \\ 
\bottomrule
\end{tabular}%

}
\caption{Comparison of quantitative results on HDRTV4K (known degradation benchmark). The first group contains CNN-based methods, while the second group includes transformer-based methods. All methods were trained using all degraded SDR versions from the HDRTV4K dataset. \textbf{Bold} and \underline{underline} indicate the best and second-best results, respectively.}
% \vspace{-3mm}
\label{tab:comparison}
\end{table*}
\subsection{Degradation-Domain Aware Controlled Mapping Network}

\subsubsection{Degradation-Aware Feature Modulation}
To handle spatially-varying SDR degradations, we design a hierarchical degradation-aware feature modulation mechanism composed of two sequential stages: a stack of $N$ Dense Controlled Mapping (DCM) modules followed by $N$ Sparse Controlled Mapping (SCM) modules. These modules inject degradation priors into the network via affine transformations over layer-normalized features, enabling both global and local adaptation to degradation patterns. Inspired by ControlNet~\cite{Zhang2023AddingCC}, we insert zero-initialized control layers before each modulation to gradually introduce prior information in a stable manner.

Each DCM module performs global, channel-wise modulation conditioned on a compact degradation embedding \( z_{\text{deg}}^g \in \mathbb{R}^{C \times 1 \times 1} \). A pair of zero-initialized \(1 \times 1\) convolutions predict affine parameters \( \gamma \) and \( \beta \), which are applied to normalized features as \( \hat{x} = \gamma \cdot x + \beta \). To enhance modulation flexibility, we extend this to degradation-aware LayerNorm, where the affine parameters are dynamically predicted from \( z_{\text{deg}}^g \). We stack $N$ such DCM modules to progressively inject global degradation priors.

After global modulation, we apply $N$ SCM modules to further refine features under spatially-varying degradations. Each SCM module utilizes a local degradation map \( z_{\text{deg}}^l \in \mathbb{R}^{C \times H \times W} \) to generate pixel-wise affine parameters. These are obtained via two parallel branches, each consisting of a shared \(1 \times 1\) convolution, a ReLU activation, and a zero-initialized \(3 \times 3\) convolution. The resulting parameters modulate the features using \( \hat{x} = \gamma \odot x + \beta \), enabling spatially adaptive transformation. Outputs from both DCM and SCM stages are fused via residual connections, forming a unified degradation-aware modulation pipeline.

\begin{table}[!t]
\centering
\renewcommand{\arraystretch}{1.15}
\setlength{\tabcolsep}{8pt}
\small
% \resizebox{0.47\textwidth}{!}{
\begin{tabular}{lcc}
\toprule
\textbf{Method} & \textbf{$\Delta E_{ITP}\downarrow$} & \textbf{HDR-VDP3$\uparrow$} \\
\midrule
HDRTVNet     & 42.4653 & 8.69 \\
FMNet        & 42.2242 & 8.75 \\
ICTCPNet     & \underline{38.1077} & 8.90 \\
LSNet        & 39.7596 & \underline{9.19} \\
AirNet       & 64.8582 & 8.48 \\
PromptIR     & 39.6666 & 9.11 \\
RAM-PromptIR & 39.4649 & 9.17 \\
RealRep & \textbf{25.9897} & \textbf{9.35} \\
\bottomrule
\end{tabular}  
% }
\caption{
Quantitative comparison using HDR-specific perceptual quality metrics.
\textbf{Bold} and \underline{underline} indicate the best and second-best results, respectively.
}
% \vspace{-2mm}
\label{tab:perceptual_metrics}
\end{table}

\subsubsection{Detail Refinement}
To enhance textures and remove residual artifacts, we append a detail refinement module consisting of stacked ResBlocks:
\begin{equation}
\resizebox{0.9\linewidth}{!}{$
I_{\text{HDR}} = F_{\text{refine}}(I_{\text{SCM}}) = \text{Conv}_{3\times3} \circ [\text{ResBlock}]^{N_r} \circ \text{Conv}_{3\times3}(I_{\text{SCM}})
$}
\end{equation}
where \( I_{\text{SCM}} \) is the output after the final SCM module, and \( N_r \) is the number of residual blocks.

\subsubsection{Training Strategy}
We adopt a two-stage training scheme. In Stage 1, contrastive learning is disabled and control layers are initialized as standard convolutions, allowing stable optimization of the encoder and modulation modules. In Stage 2, contrastive supervision is introduced via an EMA-updated negative encoder~\cite{He2019MomentumCF}, and all control layers are re-initialized as zero-convs to softly inject degradation priors.

The total loss combines pixel-level supervision and contrastive regularization:
\begin{equation}
\resizebox{0.9\linewidth}{!}{$
\mathcal{L}_{\text{total}} = \| I_{\text{HDR}}^{\text{pred}} - I_{\text{HDR}}^{\text{G.T.}} \|_1 + \lambda_l \| I_{\text{DCM}} - I_{\text{HDR}}^{\text{G.T.}} \|_1 + \lambda_{\text{contra}} \mathcal{L}_{\text{contra}}
$}
\end{equation}
where \( I_{\text{DCM}} \) denotes the output of the final DCM module. We empirically set \( \lambda_l = 0.7 \), \( \lambda_{\text{contra}} = 0.2 \).

\begin{table}[!t]
\centering

\footnotesize

\resizebox{0.47\textwidth}{!}{ % 缩放到单栏宽度
\begin{tabular}{l|cccc}
\toprule
\textbf{Metrics} & \textbf{HDRUNet} & \textbf{HDRTVNet} & \textbf{FMNet} & \textbf{ICTCPNet} \\
\midrule
PSNR            & 20.83            & \underline{23.78}     & 21.61          & 21.71 \\
SSIM            & 0.8667           & \underline{0.8824}    & 0.8609         & 0.7961 \\
\midrule
\textbf{Metrics} & \textbf{LSNet} & \textbf{PromptIR} & \textbf{RAM-PromptIR} & \textbf{Ours} \\
\midrule
PSNR            & 20.36     & 21.24          & 21.39          & \textbf{29.12} \\
SSIM            & 0.8603    & 0.7946         & 0.8089         & \textbf{0.9412} \\
\bottomrule
\end{tabular}
}
\caption{Comparison of quantitative results on HDRTV1K (unknown degradation benchmark). All methods were trained using all degraded SDR versions from the HDRTV4K dataset. \textbf{Bold} and \underline{underline} indicate the best and second-best results, respectively.}
% \vspace{-2mm}
\label{tab:hdrtv1k}
\end{table}

\begin{figure*}[tb]
    \centering
    \includegraphics[width=0.999\textwidth]{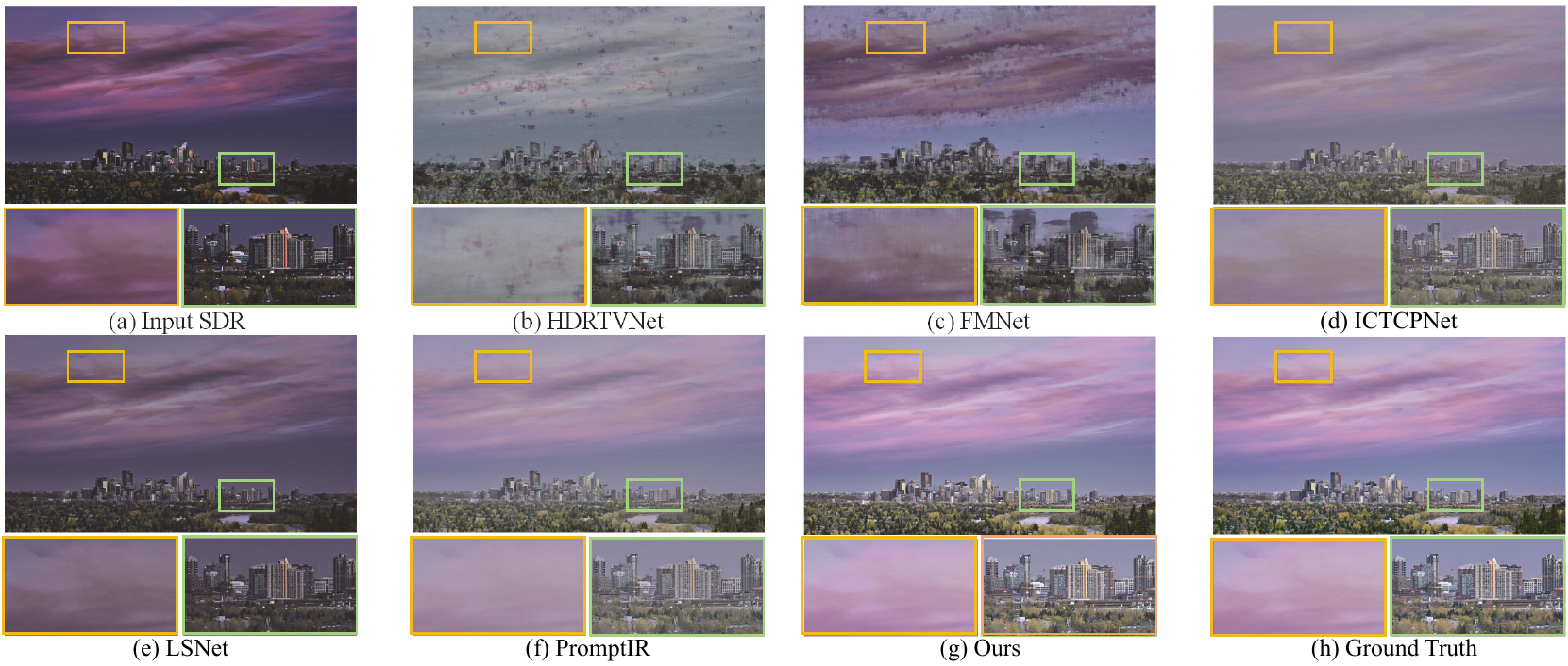}
    \caption{Visual comparison with the state-of-the-art methods. Following hdrtvdm \protect\cite{Guo2023}, the HDR images are shown in their original encoding without tone mapping to preserve highlights. Our method delivers visually superior results, with clearer highlight recovery, more vivid colors, and better structure preservation compared to prior methods. Visualization is best viewed on an HDR screen.}
    \vspace{-4mm}
    \label{Figure 6}
\end{figure*}
\section{Experiments}
\subsection{Experimental Setup}
\subsubsection{Dataset.}
We evaluate the effectiveness of our proposed framework on the HDRTV4K dataset~\cite{Guo2023}, which consists of 3878 4K HDR images encoded in BT.2020/PQ1000. All degradation types included in HDRTV4K are utilized for training and testing. The training set contains 3478 images from HDRTV4K, while the testing set comprises 400 images from HDRTV4K. Additionally, we include 117 images from the HDRTV1K dataset (BT.2020/PQ1000)~\cite{Chen2021} in the testing set to evaluate the performance on unknown degradations, resulting in a total of 517 test images. The SDR content is configured with BT.709 gamut, gamma 2.2, and a peak brightness of 100 nits.
% \vspace{-5mm}

\subsubsection{Implementation Details.}
Each multi-view encoder outputs a global vector and a local feature map with 64 and 16 channels, respectively. The mapping network consists of convolutional layers with 32 channels. We implement our framework using the PyTorch library and adopt the Adam optimizer with $\beta_1 = 0.9$, $\beta_2 = 0.99$, and a learning rate initialized to $2 \times 10^{-4}$. The learning rate is scheduled using a MultiStepLR strategy, with decay milestones at 200K and 400K iterations and a decay factor of 0.5. We employ exponential moving average (EMA) with a decay factor of 0.999 to stabilize training. The model is trained for 400K iterations in total, with the first 40K iterations designated as Stage 1. Mixed-precision training with bfloat16 is enabled to improve computational efficiency. We use a mini-batch size of 16, and the training is conducted on an NVIDIA RTX 4090 GPU, taking approximately 3 days to complete.
% \subsubsection{Comparison Methods.}

% \subsubsection{Implementation Details.}
\subsection{Experimental Results}
\subsubsection{Quantitative Results.}

We validate RealRep on the HDRTV4K (known degradation benchmark) and HDRTV1K (unknown degradation benchmark) datasets, comparing it to state-of-the-art SDR-to-HDR methods, including HDRUNet~\cite{Chen2021HDRUNetSI}, HDRTVNet~\cite{Chen2021}, FMNet~\cite{Xu2022FMNet}, ICTCPNet~\cite{Huang2023VideoIT}, LSNet~\cite{Guo2023}, PromptIR~\cite{Potlapalli2023PromptIRPF}, and RAM-PromptIR~\cite{Qin2024RestoreAW}.
To comprehensively evaluate visual quality, we report both traditional metrics such as PSNR and SSIM, as well as HDR-specific perceptual quality metrics, including HDR-VDP3 and the error-aware HDR metrics EHL, FHLP, EWG, and FWGP proposed by ~\cite{Guo2023}.  
Additionally, we evaluate perceptual color difference using $\Delta E_{\mathrm{ITP}}$, which is reported in the supplementary material.

As shown in Table~\ref{tab:comparison}, RealRep outperforms all competing methods on HDRTV4K, demonstrating its ability to handle diverse degradations and produce high-quality HDR reconstructions. This stems from its disentangled multi-view representation learning, which separates luminance and chrominance distributions, enabling robust adaptation to varying degradation characteristics. On HDRTV1K (Table \ref{tab:hdrtv1k}), RealRep excels in generalizing to unknown degradations, outperforming methods like HDRUNet, HDRTVNet, and FMNet, which suffer significant performance drops. PromptIR, while effective in simpler tasks like noise removal, struggles with complex luminance and chrominance transformations. RealRep consistently delivers superior HDR outputs, leveraging its disentangled encoder and domain-adaptive transfer framework.
\subsubsection{Qualitative Results.} 

Figure~\ref{Figure 6} compares HDR/WCG images generated by state-of-the-art methods and our proposed RealRep framework. Existing methods, such as HDRTVNet, FMNet, and ICTCPNet, struggle with diverse degradations, causing artifacts like banding, color distortion, and uneven luminance due to their reliance on fixed tone mapping strategies, which limits adaptability and leads to inconsistent performance. For instance, HDRTVNet produces overexposure and banding, FMNet shows luminance and chrominance mismatches, and ICTCPNet renders desaturated colors. In contrast, RealRep combines disentangled multi-view degradation representation learning with a domain-adaptive hierarchical transfer framework to dynamically adapt to varying degradations, producing HDR images that are natural, consistent, and artifact-free. These qualities are evident in Figure~\ref{Figure 6}, where RealRep achieves smoother luminance transitions, better highlight recovery, and accurate color distributions, surpassing other methods in handling unknown degradations and producing results closer to the ground truth.

\begin{figure}[h]
	\centering
	\includegraphics[width=0.47\textwidth]{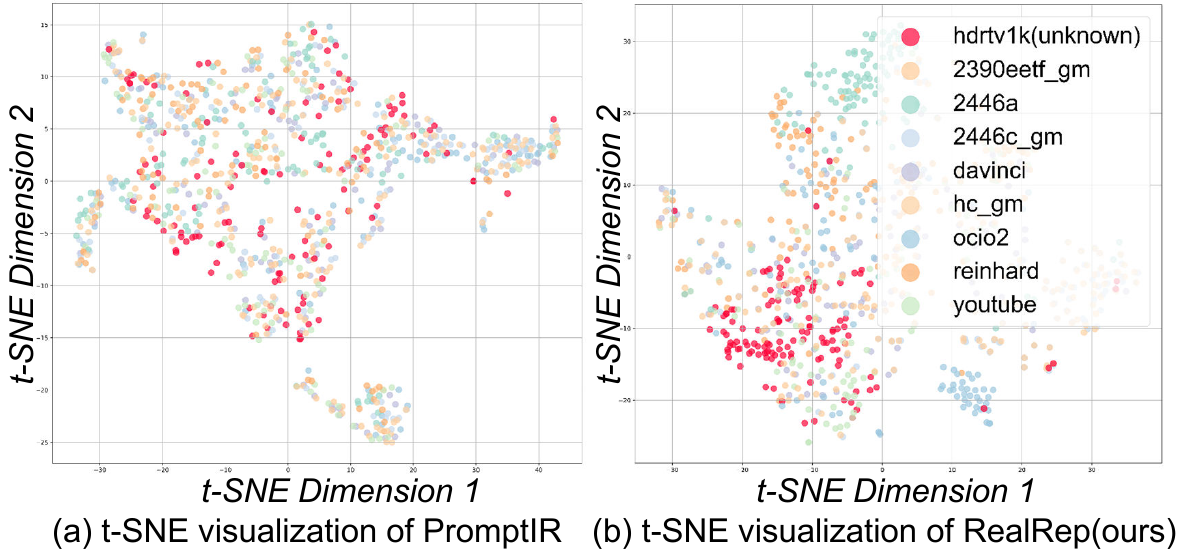}
    % \vspace{-5mm}
	\caption{Comparison of t-SNE visualizations for PromptIR (left) and our method (right). Each point represents a frame-level feature projected via PCA (16 components) and t-SNE (perplexity=50). Red points indicate unknown degradations. Our method yields a more compact and well-separated feature space, clearly distinguishing known and unknown degradations.}
    % \vspace{-2mm}
    \label{fig:tsne_visualization}
\end{figure}

\subsubsection{Joint Luma-/Chroma-aware Representation Visualization.}
To demonstrate the effectiveness of RealRep in constructing the inverse tone mapping (iTM) embedding space and its generalization to unknown degradations, we performed t-SNE visualization of the feature space. As shown in Figure~\ref{fig:tsne_visualization}, the left panel depicts the prompt representations extracted by PromptIR, while the right panel illustrates the degradation features extracted by RealRep. In PromptIR's feature space, unknown degradations (e.g., HDRTV1K) are scattered and overlap with other clusters, indicating poor generalization to unseen scenarios. In contrast, RealRep produces compact and well-separated clusters for HDRTV1K, showcasing superior generalization to unknown degradations and a more structured feature space. These findings highlight the strength of RealRep in addressing diverse real-world degradations.
\begin{table}[t]
\centering
\setlength{\tabcolsep}{3pt} % Adjust column spacing
\footnotesize % Adjust font size

\resizebox{1.0\linewidth}{!}{%
\begin{tabular}{lcccccc}
\toprule
\textbf{Configuration} & 2390eetf\_gm & 2446a & 2446c\_gm & davinci & \textbf{Avg.} \\
\midrule
Baseline              & 27.37                  & 27.72          & 25.96                & 25.67           & 25.74      \\
RealRep w/o fusion       & 32.27                  & 32.76          & 29.68                & 27.71           & 29.83      \\
RealRep w/o $z_{\text{lum}}$ & 32.96                  & 27.88          & 27.53                & 29.37           & 28.96      \\
RealRep w/o $z_{\text{chr}}$ & 32.92                  & 29.48          & 28.40                & 29.42           & 29.58      \\
RealRep w/o $\mathcal{L}_{\text{contra}}$       & 33.73        & 29.71          & 28.23                & 29.77           & 30.04      \\
RealRep & \textbf{34.29}         & \textbf{34.83} & \textbf{31.10}       & \textbf{30.17}  & \textbf{31.42} \\
\bottomrule
\end{tabular}%
}
\caption{Ablation on multi-view representations and multi-view fusion, evaluated by PSNR (dB). The final column reports the average across all eight degradation types. Due to space limitations, only four representative degradations are shown.}
% \vspace{-4mm}
\label{tab:ablation_multi_view}
\end{table}

% \vspace{-5mm}
\begin{table}[t]
\centering
\setlength{\tabcolsep}{1pt} % Adjust column spacing
% \footnotesize % Adjust font size

\resizebox{1.0\linewidth}{!}{%
\begin{tabular}{lcccccc}
\toprule
\textbf{Configuration} & 2390eetf\_gm & 2446a & 2446c\_gm & davinci & hdrtv1k & \textbf{Avg.} \\
\midrule
RealRep w/o $z_{\text{global}}$ & 24.56 & 23.43 & 28.41 & 22.41 & 25.03 & 25.00 \\
RealRep w/o $z_{\text{local}}$ & 32.95 & 34.00 & 30.90 & 27.33 & 28.42 & 30.28 \\
RealRep w/o control & 33.24 & 33.55 & 30.35 & 29.49 & 28.37 & 30.47 \\
RealRep & 34.29 & 34.83 & 31.10 & 30.17 & 29.12 & 31.16 \\
\bottomrule
\end{tabular}
}
\caption{Ablation on the degradation-aware controlled mapping module, evaluated by PSNR (dB).}
% \vspace{-4mm}
\label{tab:ablation_transfer}
\end{table}

\subsection{Ablation Studies} 

In this section, we perform ablation studies to evaluate the impact of multi-view representations and the degradation-aware controlled mapping module.

\subsubsection{Ablation of Disentangled Multi-View Representations}  
We evaluate the impact of disentangled multi-view representations through ablations on the luminance representation \( z_{\text{lum}} \), chrominance representation \( z_{\text{chr}} \), their fusion, the contrastive loss \( \mathcal{L}_{\text{contra}} \), and the fusion module. As shown in Table~\ref{tab:ablation_multi_view}, the baseline without multi-view components yields the lowest PSNR (25.74dB), confirming their necessity. Removing \( z_{\text{lum}} \) or \( z_{\text{chr}} \) reduces PSNR to 28.96dB and 29.58dB, respectively, indicating their individual contributions. Without the fusion module, PSNR drops to 29.83dB, showing its role in integrating luminance and chrominance features.
Excluding the contrastive loss \( \mathcal{L}_{\text{contra}} \) leads to a notable drop from 31.42dB to 30.04dB, highlighting the benefit of contrastive supervision. The full RealRep model achieves the highest PSNR (31.42dB), validating the joint effectiveness of all components in enhancing SDR-to-HDR performance.

\subsubsection{Ablation of Degradation-Domain Aware Controlled Mapping Network}  
We conduct ablation studies on the proposed degradation-aware controlled module, focusing on its global/local mapping components and the control mechanism. As shown in Table~\ref{tab:ablation_transfer}, removing either the global or local mapping leads to notable performance drops (25.00dB and 30.28dB average PSNR, respectively), confirming their complementary roles in modeling hierarchical domain knowledge.
The control mechanism, composed of a zero-initialized convolution and a two-stage training strategy, also proves essential. Its removal reduces average PSNR from 31.16dB to 30.78dB, and performance on HDRTV1K from 29.12dB to 28.87dB, indicating weakened generalization.
Overall, the complete RealRep model achieves the best results, demonstrating the effectiveness of the full degradation-aware controlled mapping design.

\subsubsection{Effectiveness on Unknown TM-Style SDR Inputs}
Our method is designed to generalize better to unknown tone-mapped SDR inputs.
As shown in Table~\ref{tab:tm_ablation}, our method significantly outperforms a classification-based baseline (ResNet+LSNet) on HDRTV1K, achieving higher PSNR and HDR-VDP3 scores.

\begin{table}[h]
\centering

\begin{tabular}{lcc}
\toprule
Method & PSNR & HDR-VDP3 \\
\midrule
ResNet+LSNet & 21.57 & 9.24 \\
Ours         & \textbf{29.12} & \textbf{9.60} \\
\bottomrule
\end{tabular}
\caption{Generalization to unknown TM-style SDR inputs on HDRTV1K.}
% \vspace{-4mm}
\label{tab:tm_ablation}
\end{table}

% \vspace{-5mm}
\section{Concluding Remarks}
This paper presents RealRep, a novel framework for converting SDR to HDR, designed to address real-world variations in style and degradation. By using style disentangled representation learning and a degradation-aware mapping, RealRep effectively expands luminance and color ranges. Extensive evaluations show that RealRep outperforms existing methods in visual quality and generalization.

\section*{Acknowledgements}
This work was supported in part by China Postdoctoral Science Foundation under Grant 2025M773501; in part by the Fundamental Research Funds for the Central Universities under Grant ZYTS25270.

\bibliography{aaai2026}

@article{Chen2021,
  title={A New Journey from SDRTV to HDRTV},
  author={Xiangyu Chen and Zhengwen Zhang and Jimmy S. J. Ren and Lynhoo Tian and Y. Qiao and Chao Dong},
  journal={2021 IEEE/CVF International Conference on Computer Vision (ICCV)},
  year={2021},
  pages={4480-4489},
  url={https://api.semanticscholar.org/CorpusID:237194979}
}

@inproceedings{he2022sdrtv,
  title={SDRTV-to-HDRTV via hierarchical dynamic context feature mapping},
  author={He, Gang and Xu, Kepeng and Xu, Li and Wu, Chang and Sun, Ming and Wen, Xing and Tai, Yu-Wing},
  booktitle={Proceedings of the 30th ACM international conference on multimedia},
  pages={2890--2898},
  year={2022}
}

@article{Ma2023Overview,
  author    = {Ma, S. and Gao, J. and Wang, R. and others},
  title     = {{Overview of intelligent video coding: from model-based to learning-based approaches}},
  journal   = {Visual Intelligence},
  year      = {2023},
  volume    = {1},
  articleno = {15},
  month     = {aug},
  doi       = {10.1007/s44267-023-00018-7},
  publisher = {Springer}
}

@article{xu2022sdrtv,
  title={Sdrtv-to-hdrtv conversion via spatial-temporal feature fusion},
  author={Xu, Kepeng and Xu, Li and He, Gang and Wu, Chang and Ma, Zijia and Sun, Ming and Tai, Yu-Wing},
  journal={arXiv preprint arXiv:2211.02297},
  year={2022}
}

@article{xu2023towards,
  title={Towards robust sdrtv-to-hdrtv via dual inverse degradation network},
  author={Xu, Kepeng and Xu, Li and He, Gang and Yu, Wenxin and Li, Yunsong},
  journal={arXiv e-prints},
  pages={arXiv--2307},
  year={2023}
}

@inproceedings{xu2024beyond,
  title={Beyond Alignment: Blind Video Face Restoration via Parsing-Guided Temporal-Coherent Transformer},
  author={Xu, Kepeng and Xu, Li and He, Gang and Yu, Wenxin and Li, Yunsong},
  booktitle={Proceedings of the Thirty-Third International Joint Conference on  Artificial Intelligence, $\{$IJCAI-24$\}$},
  pages={1489--1497},
  year={2024}
}

@article{xu2024beyond2,
  title={Beyond Feature Mapping GAP: Integrating Real HDRTV Priors for Superior SDRTV-to-HDRTV Conversion},
  author={Xu, Kepeng and Xu, Li and He, Gang and Zhang, Zhiqiang and Yu, Wenxin and Wang, Shihao and Zhou, Dajiang and Li, Yunsong},
  journal={arXiv preprint arXiv:2411.10775},
  year={2024}
}

@inproceedings{Kim2018,
  title={A Multi-purpose Convolutional Neural Network for Simultaneous Super-Resolution and High Dynamic Range Image Reconstruction},
  author={Soo Ye Kim and Munchurl Kim},
  booktitle={Asian Conference on Computer Vision},
  year={2018},
  url={https://api.semanticscholar.org/CorpusID:168633755}
}

@article{Kim2019Deep,
  title={Deep SR-ITM: Joint Learning of Super-Resolution and Inverse Tone-Mapping for 4K UHD HDR Applications},
  author={Soo Ye Kim and Jihyong Oh and Munchurl Kim},
  journal={2019 IEEE/CVF International Conference on Computer Vision (ICCV)},
  year={2019},
  pages={3116-3125},
  url={https://api.semanticscholar.org/CorpusID:131779280}
}

@inproceedings{Kim2019JSIGANGJ,
  title={JSI-GAN: GAN-Based Joint Super-Resolution and Inverse Tone-Mapping with Pixel-Wise Task-Specific Filters for UHD HDR Video},
  author={Soo Ye Kim and Jihyong Oh and Munchurl Kim},
  booktitle={AAAI Conference on Artificial Intelligence},
  year={2019},
  url={https://api.semanticscholar.org/CorpusID:202542696}
}

@article{Chen2021HDRUNetSI,
  title={HDRUNet: Single Image HDR Reconstruction with Denoising and Dequantization},
  author={Xiangyu Chen and Yihao Liu and Zhengwen Zhang and Y. Qiao and Chao Dong},
  journal={2021 IEEE/CVF Conference on Computer Vision and Pattern Recognition Workshops (CVPRW)},
  year={2021},
  pages={354-363},
  url={https://api.semanticscholar.org/CorpusID:235212387}
}

@article{Xu2022FMNet,
  title={FMNet: Frequency-Aware Modulation Network for SDR-to-HDR Translation},
  author={Gang Xu and Qibin Hou and Le Zhang and Ming-Ming Cheng},
  journal={Proceedings of the 30th ACM International Conference on Multimedia},
  year={2022},
  url={https://api.semanticscholar.org/CorpusID:252783139}
}

@article{Shao2022HybridCD,
  title={Hybrid Conditional Deep Inverse Tone Mapping},
  author={Tong Shao and Deming Zhai and Junjun Jiang and Xianming Liu},
  journal={Proceedings of the 30th ACM International Conference on Multimedia},
  year={2022},
  url={https://api.semanticscholar.org/CorpusID:252782392}
}

@inproceedings{Guo2023,
  author={Guo, Cheng and Fan, Leidong and Xue, Ziyu and Jiang, Xiuhua},
  booktitle={2023 IEEE/CVF Conference on Computer Vision and Pattern Recognition (CVPR)}, 
  title={Learning a Practical SDR-to-HDRTV Up-conversion using New Dataset and Degradation Models}, 
  year={2023},
  volume={},
  number={},
  pages={22231-22241},
  doi={10.1109/CVPR52729.2023.02129}
}

@inproceedings{Chen_Wang2021,  
 title={Pre-Trained Image Processing Transformer}, 
 url={http://dx.doi.org/10.1109/cvpr46437.2021.01212}, 
 DOI={10.1109/cvpr46437.2021.01212}, 
 booktitle={2021 IEEE/CVF Conference on Computer Vision and Pattern Recognition (CVPR)}, 
 author={Chen, Hanting and Wang, Yunhe and Guo, Tianyu and Xu, Chang and Deng, Yiping and Liu, Zhenhua and Ma, Siwei and Xu, Chunjing and Xu, Chao and Gao, Wen}, 
 year={2021}, 
 month={Jun}, 
 language={en-US} 
 }

@article{Li2022AllInOneIR,
  title={All-In-One Image Restoration for Unknown Corruption},
  author={Boyun Li and Xiao Liu and Peng Hu and Zhongqin Wu and Jiancheng Lv and Xiaocui Peng},
  journal={2022 IEEE/CVF Conference on Computer Vision and Pattern Recognition (CVPR)},
  year={2022},
  pages={17431-17441},
  url={https://api.semanticscholar.org/CorpusID:250551851}
}

@inproceedings{Potlapalli2023PromptIRPF,
  title={PromptIR: Prompting for All-in-One Image Restoration},
  author={Potlapalli, Vaishnav and Zamir, Syed Waqas and Khan, Salman and Khan, Fahad},
  booktitle={Thirty-seventh Conference on Neural Information Processing Systems},
  year={2023}
}

@techreport{BT2020,
  author       = {BT Series},
  title        = {Parameter values for ultra-high definition television systems for production and international programme exchange},
  institution  = {International Telecommunication Union - Telecommunication Standardization Sector (ITU-T)},
  year         = {2012},
  number       = {BT.2020},
  pages        = {1--7}
}

@techreport{SMPTE2014,
  author       = {SMPTE Standard},
  title        = {High dynamic range electro-optical transfer function of mastering reference displays},
  institution  = {Society of Motion Picture and Television Engineers (SMPTE)},
  year         = {2014},
  number       = {ST 2084},
  pages        = {11}
}

@techreport{ITU-R_BT2100-1,
  author       = {International Telecommunication Union (ITU)},
  title        = {Image Parameter Values for High Dynamic Range Television for Use in Production and International Programme Exchange},
  institution  = {International Telecommunication Union, Geneva, Switzerland},
  number       = {ITU-R BT.2100-2},
  edition      = {2},
  year         = {2018},
  month        = jul,
  pages        = {1--2}
}

@techreport{ITU-R_BT2020-2,
  author       = {International Telecommunication Union (ITU)},
  title        = {Parameter Values for Ultra-High Definition Television Systems for Production and International Programme Exchange},
  institution  = {International Telecommunication Union, Geneva, Switzerland},
  number       = {ITU-R BT.2020-2},
  edition      = {2},
  year         = {2015},
  month        = oct,
  pages        = {1--2}
}

@InProceedings{Ronneberger2015UNetCN,
author="Ronneberger, Olaf
and Fischer, Philipp
and Brox, Thomas",
title="U-Net: Convolutional Networks for Biomedical Image Segmentation",
booktitle="Medical Image Computing and Computer-Assisted Intervention -- MICCAI 2015",
year="2015",
publisher="Springer International Publishing",
address="Cham",
pages="234--241",
isbn="978-3-319-24574-4"
}

@inbook{Dong_Loy_He_Tang_2014,   title={Learning a Deep Convolutional Network for Image Super-Resolution},  url={http://dx.doi.org/10.1007/978-3-319-10593-2_13},  DOI={10.1007/978-3-319-10593-2_13},  booktitle={Computer Vision – ECCV 2014,Lecture Notes in Computer Science},  author={Dong, Chao and Loy, Chen Change and He, Kaiming and Tang, Xiaoou},  year={2014},  month={Jan},  pages={184–199},  language={en-US}  }

@inproceedings{Kim_Lee_Lee_2016,   title={Accurate Image Super-Resolution Using Very Deep Convolutional Networks},  url={http://dx.doi.org/10.1109/cvpr.2016.182},  DOI={10.1109/cvpr.2016.182},  booktitle={2016 IEEE Conference on Computer Vision and Pattern Recognition (CVPR)},  author={Kim, Jiwon and Lee, Jung Kwon and Lee, Kyoung Mu},  year={2016},  month={Jun},  language={en-US}  }

@inbook{Ahn_Kang_Sohn_2018,   title={Fast, Accurate, and Lightweight Super-Resolution with Cascading Residual Network},  url={http://dx.doi.org/10.1007/978-3-030-01249-6_16},  DOI={10.1007/978-3-030-01249-6_16},  booktitle={Computer Vision – ECCV 2018,Lecture Notes in Computer Science},  author={Ahn, Namhyuk and Kang, Byungkon and Sohn, Kyung-Ah},  year={2018},  month={Jan},  pages={256–272},  language={en-US}  }

@inproceedings{Lim_Son_Kim_Nah_Lee_2017,   title={Enhanced Deep Residual Networks for Single Image Super-Resolution},  url={http://dx.doi.org/10.1109/cvprw.2017.151},  DOI={10.1109/cvprw.2017.151},  booktitle={2017 IEEE Conference on Computer Vision and Pattern Recognition Workshops (CVPRW)},  author={Lim, Bee and Son, Sanghyun and Kim, Heewon and Nah, Seungjun and Lee, Kyoung Mu},  year={2017},  month={Jul},  language={en-US}  }

@article{Tian_Xu_Li_Zuo_Fei_Liu_2020,   title={Attention-guided CNN for image denoising},  url={http://dx.doi.org/10.1016/j.neunet.2019.12.024},  DOI={10.1016/j.neunet.2019.12.024},  journal={Neural Networks},  author={Tian, Chunwei and Xu, Yong and Li, Zuoyong and Zuo, Wangmeng and Fei, Lunke and Liu, Hong},  year={2020},  month={Apr},  pages={117–129},  language={en-US}  }

@article{Yang_Liu_Yang_Guo_2019,   title={Scale-Free Single Image Deraining Via Visibility-Enhanced Recurrent Wavelet Learning},  url={http://dx.doi.org/10.1109/tip.2019.2892685},  DOI={10.1109/tip.2019.2892685},  journal={IEEE Transactions on Image Processing},  author={Yang, Wenhan and Liu, Jiaying and Yang, Shuai and Guo, Zongming},  year={2019},  month={Jun},  pages={2948–2961},  language={en-US}  }

@article{Li_Peng_Wang_Xu_Feng_2017,   title={An All-in-One Network for Dehazing and Beyond},  journal={Cornell University - arXiv,Cornell University - arXiv},  author={Li, Boyi and Peng, Xiulian and Wang, Zhangyang and Xu, Jizheng and Feng, Dan},  year={2017},  month={Jul},  language={en-US}  }

@article{Eboli_Sun_Ponce_2020,   title={End-to-end Interpretable Learning of Non-blind Image Deblurring},  journal={Le Centre pour la Communication Scientifique Directe - HAL - Diderot,Le Centre pour la Communication Scientifique Directe - HAL - Diderot},  author={Eboli, Thomas and Sun, Jian and Ponce, Jean},  year={2020},  month={Aug},  language={en-US}  }

@article{Huang2023VideoIT,
  title={Video Inverse Tone Mapping Network with Luma and Chroma Mapping},
  author={Peihuan Huang and Gaofeng Cao and Fei Zhou and Guoping Qiu},
  journal={Proceedings of the 31st ACM International Conference on Multimedia},
  year={2023},
  url={https://api.semanticscholar.org/CorpusID:264492628}
}

@article{Zhang2023AddingCC,
  title={Adding Conditional Control to Text-to-Image Diffusion Models},
  author={Lvmin Zhang and Anyi Rao and Maneesh Agrawala},
  journal={2023 IEEE/CVF International Conference on Computer Vision (ICCV)},
  year={2023},
  pages={3813-3824},
  url={https://api.semanticscholar.org/CorpusID:256827727}
}

@inproceedings{Qin2024RestoreAW,
  title={Restore Anything with Masks: Leveraging Mask Image Modeling for Blind All-in-One Image Restoration},
  author={Chuntao Qin and Ruiqi Wu and Zikun Liu and Xin Lin and Chun-Le Guo and Hyun Hee Park and Chongyi Li},
  booktitle={European Conference on Computer Vision},
  year={2024},
  url={https://api.semanticscholar.org/CorpusID:272987034}
}

@article{He2019MomentumCF,
  title={Momentum Contrast for Unsupervised Visual Representation Learning},
  author={Kaiming He and Haoqi Fan and Yuxin Wu and Saining Xie and Ross B. Girshick},
  journal={2020 IEEE/CVF Conference on Computer Vision and Pattern Recognition (CVPR)},
  year={2019},
  pages={9726-9735},
  url={https://api.semanticscholar.org/CorpusID:207930212}
}

@inproceedings{MaIR,
  title={MaIR: A Locality- and Continuity-Preserving Mamba for Image Restoration},
  author={Li, Boyun and Zhao, Haiyu and Wang, Wenxin and Hu, Peng and Gou, Yuanbiao and Peng, Xi},
  booktitle = {IEEE Conference on Computer Vision and Pattern Recognition},
  year = {2025},
  address = {Nashville, TN},
  month = jun
}

\end{document}